\documentclass[runningheads,10pt]{llncs}

\usepackage[english]{babel}
\usepackage{url}
\usepackage{amsmath}
\usepackage{amssymb}
\usepackage{amsfonts}

\begin{document}

\title{On Building a Knowledge Base for Stability Theory
\thanks{The final publication of this paper is available at www.springerlink.com.}}

\author{Agnieszka Rowinska-Schwarzweller\inst{1}
\and
Christoph Schwarzweller\inst{2}}

\authorrunning{A. Rowinska-Schwarzweller and C. Schwarzweller}

\institute{Chair of Display Technology, University of Stuttgart\\
           Allmandring 3b, 70569 Stuttgart, Germany \\
              \email{schwarzweller@neostrada.pl}
\and
Department of Computer Science, University of Gda\'{n}sk\\
           ul. Wita Stwosza 57, 80-952 Gda\'nsk, Poland\\
            \email{schwarzw@inf.ug.edu.pl}}

\maketitle

\begin{abstract}
A lot of mathematical knowledge has been formalized and stored in repositories by now:
different mathematical theorems and theories have been taken into consideration and
included in mathematical repositories. Applications more distant from pure mathematics, 
however --- though based on these theories --- often need more detailed knowledge about
the underlying theories.
In this paper we present an example Mizar formalization from the area of electrical 
engineering focusing on stability theory which is based on complex analysis. We discuss 
what kind of special knowledge is necessary here and which amount of this knowledge is 
included in existing repositories.
\end{abstract}

\section{Introduction}

The aim of mathematical knowledge management is to provide both tools 
and infrastructure supporting the organization, development,
and teaching of mathematics with the help of effective up-to-date
computer technologies.
To achieve this ambitious goal it should be taken into account that the predominant part
of potential users will not be professional mathematicians themselves, 
but rather scientists or teachers that apply mathematics in their special domain.
To attract this group of people it is essential that our repositories provide
a sufficient knowledge base for those domains.
We are interested in how far existing mathematical repositories are
from meeting this precondition yet, or in other words:
How big is the gap between the knowledge already included in repositories
and the knowledge necessary for particular applications?

This problem, however, concerns not only the simple question how much knowledge of a
domain is available in a repository.
We believe, that in order to measure this gap, it is equally important to consider the basic conditions for
a successful formalization of applications on top of existing knowledge, that is
on top of a mathematical repository:
The more easy such a formalization is, the more attractive is a mathematical repository.
To describe attractiveness of a repository for an application
one can identify three major points:
\begin{itemize}
\item[1.] Amount of knowledge \\
          This is the obvious question, how much knowledge of a particular
          domain already has been formalized and included in the repository.
          Basically, the more knowledge of a domain is included the more attractive
          a repository is for applications.\\
\item[2.] Representation of knowledge\\
          This concerns the question of how the knowledge has been defined and
          formalized: Often mathematicians use more abstract constructions than
          necessary --- and attractive --- for applications. 
          An example is the construction of rational functions from polynomials.\\
\item[3.] Applicability of knowledge\\
          This point deals with both how the knowledge of a domain is organized in a repository
          and the question of how easy it is to adapt available knowledge to
          one's own purposes.
\end{itemize}

In this paper we focus on electrical engineering, in particular on network stability \cite{unb}.
Network theory deals with the mathematical description, analysis and
synthesis of electrical (e.g. continuous, time-discrete or digital) networks.
For a reliable application such systems have to be (input/output-) stable, that is
for an arbitrary bounded input the output have to be bounded again.
In practice it is impossible to verify responses for all input signals.
In this situation there is, however, a number of theorems permitting easier
methods to decide whether a network is stable\cite{unb}.
We shall introduce the mathematical fundamentals and prequisites of one example
theorem and present a Mizar formalization of this theorem.
After that we discus our formalization in the spirit of the three points from above:
How far is the Mizar sytem from providing a suitable mathematical repository
for applications in stability theory?

\section{Networks and their Stability} \label{nets}

As mentioned in the introduction the (input/output-) stability of networks is one of the main issues
when dealing with the analysis and design of electrical circuits and systems.
In the following we briefly review definitions and properties of electrical systems
necessary to understand the rest of the paper.
In electrical engineering stability applies to the input/output behaviour
of networks (see figure 1).
For (time-) continuous systems one finds the following definition.
For discrete systems an analogous definition is used.
\begin{definition} \rm (\cite{unb})\\
A continuous system is (BIBO-)\footnote{BIBO stands for Bounded Input Bounded Output.} stable, if and only if 
each bounded input signal $x(t)$ results in a bounded output signal $y(t)$.
\end{definition}
Physically realizable, linear time-invariant systems (LTI systems) can be described
by a set of differential equations \cite{unb}.
The behaviour of a LTI system then is completely characterized by its impulse response 
$h(t)$.\footnote{$h(t)$ is the output of the system, when the input is the
Dirac delta function $\delta(t)$.}
If the impulse response of auch a system is known, the relation between the input $x(t)$
and the output $y(t)$ is given by the convolution integral
\begin{eqnarray}
y(t) \ = \ \int \limits _{-\infty}^{\infty} x(\tau) h(t-\tau) \mbox{d}\tau .
\end{eqnarray}
Furthermore, a LTI system is stable, if and only if its impulse response $h(t)$
is absolute integrable, that is there exists a constant $K$ such that
\begin{eqnarray}
\int \limits_{-\infty}^{\infty} \left | h(\tau) \right | \mbox{d}\tau \ \leq \ K \ < \ \infty .
\end{eqnarray}
In network and filter analysis and design, however, one commonly employs the frequency domain
rather than the time domain.
To this end the system is described based on its transfer function $H(s)$.
In case the Laplace transformation is used we have\footnote{Note that 
this is a generalization of the continuous-time Fourier transformation.}
\begin{eqnarray}
H(s) \ = \ \int \limits_{-\infty}^{\infty} h(t)e^{-st} \mbox{d}t . \\ \nonumber 
\end{eqnarray}
where $s = \sigma + j\omega$ is a complex variable with $\Re\{s\} = \sigma$ and
$\Im\{s\} = \omega$.


\begin{center}
\begin{figure}[h]
\begin{picture}(70,40)(-80,0)

\unitlength1mm
\thicklines
\put(17,0){\line(1,0){30}}
\put(17,0){\line(0,1){20}}
\put(47,0){\line(0,1){20}}
\put(17,20){\line(1,0){30}}
\put(47,10){\vector(1,0){10}}
\put(7,10){\vector(1,0){10}}

\put(6,10){\circle{2}}
\put(58,10){\circle{2}}

\put(0,10){\makebox(0,0){$x(t)$}}
\put(64,10){\makebox(0,0){$y(t)$}}

\put(32,4){\makebox(0,0){$H(s)$}}
\put(32,10){\makebox(0,0){$h(t)$}}
\put(32,16){\makebox(0,0){System}}

\put(0,-10){\makebox(66,0){Figure 1: LTI system with one input $x(t)$ and one output $y(t)$}}
\end{picture}
\end{figure}
\end{center}

\vspace{0,2cm}

The evaluation of $H(s)$ for $s = j\omega$ --- in case of convergence\footnote{In this case
$H(j\omega)$ equals the Fourier transform.} --- enables the
qualitative understanding of how the system handles and selects various frequencies $\omega$, so for example
whether the system describes a high-pass filter, low-pass filter, etc.
Now the necessary condition to demonstrate the stability of LTI systems in the frequency
domain reduces to show, that the $j\omega$-axis lies in the Laplace transformation's
region of convergence (ROC). 

For physically realizable LTI systems, such as the class of networks with constant
and concentrated parameters, $H(s)$ is given in form of a 
rational function with real coefficients, that is 
\begin{eqnarray}
H(s) \ = \ \frac{a_n s^n + \ldots + a_0}{b_m s^m + \ldots + b_0}, \ \ \ a_i,b_i \in \mathbb{R}.
\end{eqnarray}
In this case the region of convergence can be described by the roots of the denominator polynomial:
If $s_i = \sigma_i + j\omega_i$ for $i = 1,\ldots m$ are the roots of $b_m s^m + \ldots + b_0$,
the region of convergence is given by
\[ \Re \{ s \} \ > \max \{ \sigma_i,\ i = 1, \ldots m \} .\]
To check stability it is therefore sufficient, to show that the real part  $\Re \{ s \}$
of all poles of $H(s)$ is smaller then $0$.
The denominator of $H(s)$ is thus a so-called Hurwitz polynomial.

Note that
the stability problem for discrete-time signals and systems can be analized with the same
approach. 
For a given discrete-time transfer function $H(z)$ in the $Z$- domain, it has to be checked whether 
the unit circle is contained in the region of convergence.
Hence for all poles $z_i$ of $H(z)$ we must have $|z_i| < 1$.
Using bilinear transformations \cite{oppscha}
\begin{eqnarray}
z \ := \ \frac{1+s}{1-s}.
\end{eqnarray}
it is thus sufficient to check whether the denominator of
\begin{eqnarray}
\left. H(z) \right | _{z := \frac{1+s}{1-s}}
\end{eqnarray}
is a Hurwitz polynomial.

The practical proof of stability of high-precision filters, however, turns out to be
very hard.
In practical applications the poles of concern 
are usually close to the axis $s = j\omega$ or the unit circle $|z| = e^{j\omega}$ respectively.
Thus numerical determination of the poles is highly error-pruning due to its
rounding effects.
In digital signal processing in addition degrees of transfer functions tend to be very high,
for example 128 and higher in communication networks.

An interesting and in practice often used method to check the stability
of a given network is based on the following theorem.\\

\noindent
{\bf Theorem 1.} (\cite{unb}) \\
Let $f(x)$ be a real polynomial with degree $n \geq 1$. Furthermore let all
coefficients of $f(x)$ be greater than $0$. 
Let $f_e(x)$ and  $f_o(x)$ denote the even part resp. the odd part of $f(x)$.
Assume further that \[ Z(x) \ = \ \frac{f_e(x)}{f_o(x)} \]
or the reciprocal of $Z(x)$ is a reactance one-port function of degree $n$.
Then $f(x)$ is a Hurwitz polynomial.  \\


The concept of reactance one-port function stems from electrical network theory:
In arbitrary passive (that is RLC-) networks we find the following
relations between the complex voltage $U_{\nu}(s)$ and the complex current $I_{\nu}(s)$:

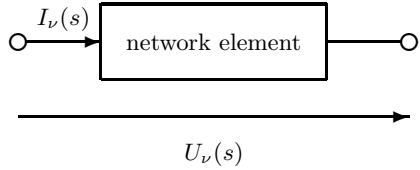
\begin{center}
\begin{figure}[h]
\begin{picture}(370,30)(-80,0)

\unitlength1mm
\thicklines
\put(17,5){\line(1,0){30}}
\put(17,5){\line(0,1){10}}
\put(47,5){\line(0,1){10}}
\put(17,15){\line(1,0){30}}
\put(47,10){\line(1,0){10}}
\put(7,10){\vector(1,0){10}}
\put(7,13){\makebox(10,0){$I_{\nu}(s)$}}

\put(6,10){\circle{2}}
\put(58,10){\circle{2}}

\put(6,0){\vector(1,0){52}}
\put(6,-5){\makebox(52,0){$U_{\nu}(s)$}}


\put(32,10){\makebox(0,0){network element}}

\end{picture}
\end{figure}
\end{center}
\vspace{-0,2cm}
\[\begin{array}{lclll}
U_{r}(s) & = & R_r \cdot I_{r}(s) & & \mbox{for a resistor } R_r \\ \\
U_{l}(s) & = & s \cdot L_l \cdot I_{l}(s) & & \mbox{for an inductance } L_l \\ \\
U_{k}(s) & = & \frac{1}{s\cdot C_k} \cdot  I_{k}(s) & & \mbox{for a capacity } C_k \\ \\
\end{array}\]

An impedance (complex resistor) or admittance (complex conductance) composed of
network elements R, L and C only is called a (RLC-) one-port function, 
an impedance or admittance composed of
network elements L and C only is called a reactance one-port function.
Conversely, for every one-port function $Z(s)$ there exists at least one one-port, whose impedance or
admittance is equal to $Z(s)$:

Hence, theorem 1 reduces stability checking to the considerable easier task to synthesize a one-port
solely using inductors (L) and capacitors (C),
that is to synthesize a reactance one-port.
To this end there exist easy procedures like for example Routh's method
to construct a chain one-port \cite{unb}.\\

It turns out that one-port functions $Z(s)$ are exactly the real positive rational functions.
For a real function we have that
for real $s$ also $Z(s)$ is real\footnote{This condition implies that the coefficients
of $Z(s)$ are real. In network theory, however, this definition is used.}, and a positive function
means that $\Re\{s\} > 0$ implies $\Re\{Z(s)\} > 0$.
A reactance one-port function is a one-port function, that is in addition odd.
The property of being positive is closely connected to Hurwitz polynomials:\\

\noindent
{\bf Theorem 2.} (\cite{unb}) \\
Let $f(x)$ be a real polynomial with degree $n \geq 1$. Furthermore let all
coefficients of $f(x)$ be greater than $0$. 
Let $f_e(x)$ and  $f_o(x)$ denote the even part resp. the odd part of $f(x)$.
Assume that$f_e(x)$ and  $f_o(x)$ have no common roots and that
$Z(x) = f_e(x)/f_o(x)$ is positive. Then
\begin{itemize}
\item[(i)] $\Re\{Z(x)\} \geq 0$ for all $x$ with $\Re\{x\} = 0$
\item[(ii)] $f_e(x) + f_o(x)$ is a Hurwitz polynomial.\\
\end{itemize}

In section \ref{routhform} we will see that this theorem is also the key to prove
that stability checking can be reduced to synthesizing LC-one-ports,
in other words to prove theorem 1 from above.

\section{The Mizar System}

The logical basis of Mizar \cite{mizar,HomePage} is classical first order logic
extended, however, with so-called schemes. Schemes introduce free second order variables, 
in this way enabling among others the definition of induction schemes.
In addition Mizar objects are typed, the types forming a hierarchy with the fundamental
type \verb@set@. The user can introduce new (sub)types describing
mathematical objects such as groups, fields, vector spaces or polynomials
over rings or fields. To this end the Mizar language provides a powerful
typing mechanism based on adjective subtypes \cite{miztypes}.

The current development of Mizar relies on 
Tarski-Grothen\-dieck set theory --- a variant of Zermelo Fraenkel set theory
using Tarski's axiom on arbitrarily large, strongly inaccessible cardinals
\cite{Tarski} which can be used to prove the axiom of choice  ---, though 
in principle the Mizar language can be used with other axiom systems also.
Mizar proofs are written in natural deduction style as presented in
the calculus of \cite{jas}. The rules of the calculus are connected with 
corresponding (English) natural language phrases so that the Mizar language
is close to the one used in mathematical textbooks. The Mizar proof checker verifies
the individual proof steps using the notion of obvious inferences \cite{obvious}
to shorten the rather long proofs of pure natural deduction.

Mizar objects are typed, the types forming a hierarchy with the fundamental
type \verb@set@ \cite{miztypes}.
New types are constructed using type constructors called modes. 
Modes can be decorated with adjectives --- given by so-called
attribute definitions --- in this way extending the type hierarchy:
For example, given the mode \verb@Ring@ and an attribute \verb@commutative@
a new mode \verb@commutative Ring@ can be constructed, which obeys all the
properties given by the mode \verb@Ring@ plus the ones stated by the
attribute \verb@commutative@.
Furthermore, a variable of type \verb@commutative Ring@ then is also of type 
\verb@Ring@, which implies that all notions defined for \verb@Ring@
are available for \verb@commutative Ring@. In addition all theorems proved for 
type \verb@Ring@
are applicable for objects of type \verb@commutative Ring@; indeed the Mizar
checker itself infers subtype relations in order to check whether notions and theorems
are applicable for a given type.


\section{Mizar Formalization of the Theorem} \label{formalization}

\subsection{Some Preliminaries About Rational Functions} \label{prel}

Although the theory of polynomials in Mizar is rather well developed,
rational functions have not been defined yet.
Rational functions can --- analogously to polynomials --- be defined over arbitrary
fields: Rational functions are simply pairs of polynomials whose second component is not the zero 
polynomial.\footnote{Of course rational functions can be
introduced "more algebraically" as the quotient field of a polynomial ring.
Here we decided to use pairs to concentrate on application issues;
see the discussion in section \ref{lessons}.}
These can be easily introduced as a Mizar type \verb@Rational_function of L@, where \verb@L@ is
the underlying coefficient domain.

{\small\begin{verbatim}
definition
let L be non trivial multLoopStr_0;
mode rational_function of L means
  ex p1 being Polynomial of L st
  ex p2 being non zero Polynomial of L st it = [p1,p2];
end;
\end{verbatim}}

In Mizar the result types of the pair constructor \verb@[ , ]@ and the projections \verb@`1@ and \verb@`2@,
that in the original definition are simply \verb@set@, then can be modified into
\verb@Rational_function@ and \verb@(non zero) Polynomial@ respectively.
In addition one can introduce the usual functions \verb@num@ and \verb@denom@ as
synonyms for the corresponding projections. \\

{\small\begin{verbatim}
definition
let L be non trivial multLoopStr_0;
let p1 be Polynomial of L;
let p2 be non zero Polynomial of L;
redefine func [p1,p2] -> rational_function of L;
end;

definition
let L be non trivial multLoopStr_0;
let z be rational_function of L;
redefine func z`1 -> Polynomial of L;
redefine func z`2 -> non zero Polynomial of L;
end;

notation
let L be non trivial multLoopStr_0;
let z be rational_function of L;
synonym num(z) for z`1;
synonym denom(z) for z`2;
end;
\end{verbatim}}

Now that \verb@num@ and \verb@denom@ --- applied to rational functions --- have result type \verb@Polynomial@,
operations for rational functions can straightforwardly be defined
by employing the corresponding functions for polynomials. So, for example, the
evaluation of rational functions can be defined using evaluation of polynomials
and the division operator \verb@/@ defined for arbitrary fields \verb@L@.

{\small\begin{verbatim}
definition
let L be Field;
let z be rational_function of L;
let x be Element of L;
func eval(z,x) -> Element of L equals
  eval(num(z),x) / eval(denom(z),x);
end;
\end{verbatim}}

Note that according to the definition of \verb@eval@ for polynomials
the type of the first argument --- that is of \verb@num(z)@ and \verb@denom(z)@ --- has to be \verb@Polynomial@.
This is ensured by the redefinitions from above, which in this sense allow
for reusing the operations defined for polynomials in the case of rational functions.
Other necessary operations for rational functions such as the degree or arithmetic operations
can be defined the same way.

\subsection{The Theorem}\label{routhform}

Using the general Mizar theory of polynomials and rational functions for our purposes, that is for
complex numbers, is straightforward.
We just instantiate the parameter {\tt L} describing the coefficient domain with
the field of complex numbers \verb@F_Complex@ from \cite{complexf}.
So an object of type
\[ \verb@rational_Function of F_Complex@ \]
combines the theory of rational functions with the one of complex numbers.

Further properties necessary to state the main theorem are introduced by defining appropriate 
attributes for polynomials and rational functions resp.
Note that these definitions apply to polynomials and rational functions over the complex
numbers only given by the instantiated Mizar types mentioned above.

{\small\begin{verbatim}
definition
let p be Polynomial of F_Complex;
attr p is real means
  for i being Element of NAT holds p.i is real number;
end;

definition
let p be rational_Function of F_Complex;
attr Z is positive means
  for x being Element of F_Complex 
  holds Re(x) > 0 implies Re(eval(Z,x)) > 0;
end;
\end{verbatim}}

Using these attributes --- and the attribute {\tt odd} describing odd functions ---
we can then introduce one-ports and reactance one-ports
in Mizar by the following mode definitions.\\

{\small\begin{verbatim}
definition
mode one_port_function is real positive rational_function of F_Complex;
mode reactance_one_port_function is
                      odd real positive rational_function of F_Complex;
end;
\end{verbatim}}

We also needed to define the odd and the even part of a polynomial \verb@f@.
This is accomplished by two Mizar functors \verb@even_part(f)@ and
\verb@odd_part(f)@, which however can be defined
straightforwardly.
Finally we formalize the condition from the theorem, that all the coefficients of the given
polynomial $f$ should be greater than $0$ as usual as a Mizar attribute:

{\small\begin{verbatim}
definition
let f be real Polynomial of F_Complex;
attr f is with_positive_coefficients means
  for i being Element of NAT st i <= deg p holds p.i > 0;
end;
\end{verbatim}}

Note that for a real polynomial $f$ with positive coefficients and deg$(f) \geq 1$ both
the even and the odd part of $f$ are not $0$, hence both can appear as the denominator of
a rational function.
Thus prepared we can state theorem 1 from section \ref{nets} in Mizar.
Note again that due to the redefinitions of section
\ref{prel} the functor \verb@[ , ]@ returns a rational function.

{\small\begin{verbatim}
theorem
for p be non constant with_positive_coefficients 
        (real Polynomial of F_Complex)
st [even_part(p),odd_part(p)] is reactance_one_port_function &
   degree([even_part(p),odd_part(p)]) = degree p
holds p is Hurwitz;
\end{verbatim}}

The proof of the theorem, as already indicated, basically relies
on theorem 2 from section \ref{nets}, which connects rational
functions with the property of being a Hurwitz polynomial.
Based on our development from above one can formulate this theorem as follows.

{\small\begin{verbatim}
theorem
for p be non constant with_positive_coefficients 
        (real Polynomial of F_Complex)
st [even_part(p),odd_part(p)] is positive & 
   even_part(p),odd_part(p) have_no_common_roots
holds (for x being Element of F_Complex 
       st Re(x) = 0 & eval(odd_part(p),x) <> 0 
       holds Re(eval([even_part(p),odd_part(p)],x)) >= 0) &
      even_part(p) + odd_part(p) is Hurwitz;
\end{verbatim}}

The corresponding Mizar proof is rather technical.
The basic idea consists of considering in addition to
\begin{eqnarray}
 Z(x) \ = \  \frac{f_e(x)}{f_o(x)}  
\end{eqnarray}
the rational function
\begin{eqnarray} \label{w}
 W(x) \ = \  \frac{Z(x)-1}{Z(x)+1} 
      & = &  \frac{\mbox{num}(Z)(x) - \mbox{denom}(Z)(x)}
                  {\mbox{num}(Z)(x) + \mbox{denom}(Z)(x)}. 
\end{eqnarray}
and to analyze the absolute values $|W(x)|$.
If $Z(x)$ is positive, then $|W(x)| \leq 1$ for all $x$ with
$\Re(x) \geq 0$, which implies
that $W(x)$ has no poles for $\Re(x)\geq 0$.
Thus the denominator polynomial can have roots only for $\Re(x) < 0$,
so $\mbox{num}(Z)(x) + \mbox{denom}(Z)(x)$ is a Hurwitz polynomial.

The main theorem now easily follows from theorem 2, because the degree condition
implies that \verb@even_part(p)@ and \verb@odd_part(p)@ have no common roots.

\section{Discussion --- Lessons Learned} \label{lessons}

In the following we discuss our formalization from the last section
with respect to the tree criteria developed in the introduction.
Though restricted to Mizar we claim that the situation in other
repositories is similar, so that most of our results hold in a more general
context also.

In \cite{schur1} we already presented a Mizar formalization of Schur's theorem,
another helpful criterion for stability checking.
Based on polynomials only its formalization was rather harmless.
The only missing point that caused some work was division
of polynomials.
However, as we will see, stability checking in general needs definitely
more extension than in this case.

\subsection{Amount of Knowledge}

Complex numbers and polynomials (over arbitrary rings) are included in MML.
A lot of theorems have been proved here, so that almost all we needed could
be found in the repository.
Interestingly rational functions --- a rather basic structure --- had not been defined, yet.
The reason might be that rational functions are mathematically rather simple
and this is the first time that a theorem relying on rational functions has been
formalized.

Though even and odd functions were already included in MML, the even and
odd part of a polynomial was not.
This, however, comes with no surprise, just because these polynomial operations
are rather seldom used.
We hence had to prove a number of theorems dealing with these polynomials,
most of them however being elementary like for example

{\small\begin{verbatim}
theorem
for p being real Polynomial of F_Complex
for x being Element of F_Complex st Re(x) = 0
holds Re(eval(odd_part(p),x)) = 0;
\end{verbatim}}

Summarizing, besides the lack of rational functions, MML provides the amount
of knowledge for stability checking one could expected.

\subsection{Representation of Knowledge}

The construction of rational functions can be performed in different ways.
Of course, one can define rational functions as pairs of polynomials.
On the other hand there is the possibility to construct (the field of)
rational functions as the completion of polynomial rings.
Though the second version is mathematically more challenging we decided
to use pairs. 
We wanted to emphasize the contribution to applications by concentrating
on prior knowledge of potential users:
Electrical engineers are probably not interested in (working with!)
abstract algebra, their interests and needs are different.

In the same context there is another representational problem:
In MML we find both the complex numbers and the field of complex numbers.
Not a problem in itself, this may cause some confusion when searching
for notions and theorems:
The functors {\t Re} and {\tt Im} giving the real and imaginary part, for exmaple, are defined
for complex numbers only, thus --- theoretically --- not applicable to elements of a field.
In Mizar, however, this is not necessarily the case:
Using a special registration --- {\tt identify} \cite{ident} --- the user can
identify terms and operations from different structures,
here complex numbers with elements of the field of complex numbers:

{\small\begin{verbatim}
registration
  let a,b be complex number;
  let x,y be Element of F_Complex;
  identify x+y with a+b when x=a, y=b;
  identify x*y with a*b when x=a, y=b;
end;
\end{verbatim}}

In effect, after this registration functors {\t Re} and {\tt Im} are applicable to 
elements of the field of complex numbers.

In general, different views on mathematical objects --- here, complex numbers
as numbers or elements of a field --- have to be handled carefully in mathematical
repositories in order to not confuse possible users.
Even the rudimentary difference between a polnomial and its polynomial function
can lead to surprises and incomprehension for people not familiar with
the formal treatment of mathematics in repositories.

\subsection{Applicability of Knowledge}

As we have already seen, the adaption of general knowledge in MML to special cases
is straightforward:
One just instantiates parameters describing the general domain with
the special one, so for example
\verb@Polynomial of F_Complex@
for polynomials over the (field of) complex numbers.

This, on the other hands, means that to work with such instantiations the user
has to apply theorems about the general structure.
Though highly desirable from the mathematical point of view,
it is not clear whether this is really convenient for application users:
To work in the special field of complex numbers, for example, then means to search for
helpful theorems in the theory of fields, rings or even groups and semigroups.
Maybe here a search tool that generates and collects theorems for
special instances of theories would be a reasonable help.

The organization of MML is mainly by articles in which authors prove
not only their main theorems, but also whatever is necessary and not found
in MML.
As a consequence theorems of the same topic, e.g. polynomials, can be spread
over the repository.
A step to overcome this shortcoming is an ongoing project called  Mizar encyclopedia building
articles with monographic character whose contents is semi-automatically
extracted from contributed Mizar articles.
Unfortunately polynomials have not been considered in this project, yet.

Summarizing the Mizar system though flexible in order to support special
applications lacks an organization of its corresponding repository
to support application users in their efforts.

\section{Conclusions} \label{concl}

We have presented a Mizar formalization of a theorem for stability checking and
have discussed how the knowledge contained in MML supported the process
from an application user's point of view.
Here we want to emphasize two points.

First, when building a knowledge base for an application area,
it is hardly foreseeable what knowledge is necessary.
We have seen that the formalization of Schur's theorem went through without
major problems, while the present theorem caused definitely more work
and preparation.
Furthermore, there are theorems on stability checking using even involved
mathematical techniques such as, e.g., analytic functions and the
maximum principle.

Second, attractiveness of mathematical repositories does not only
depend on the amount of knowledge included.
Equally important are a clear representation and organization of knowledge in the sense
that it stays familiar for users outside the mathematical community.

Consequently is it essential to communicate with experts from the
application area.
If we want our repositories to be widely used we have both
to provide a reasonable knowledge base and to take care of the fact
that application users might represent mathematical knowledge in a 
different way we are used to.
\\


\begin{thebibliography}{ABC99a}

\bibitem[Byl90]{complex} C. Byli\'{n}ski,
      The Complex Numbers;
      Formalized Mathematics, 1(3), pp.~507--513, 1990.

\bibitem[Ban03]{miztypes}
  G. Bancerek, On the Structure of Mizar Types; 
  in: H. Geuvers and F. Kamareddine (eds.),
  Proc. of MLC 2003, ENTCS 85(7), 2003.

\bibitem[Dav81]{obvious}
  M. Davies, Obvious Logical Inferences;
  in: Proceedings of the 7th International Joint Conference on 
  Artificial Intelligence, pp. 530-531, 1981.

\bibitem[DeB87]{DeBruijn} N.G. de~Bruijn,
  The Mathematical Vernacular, a language for mathematics
  with typed sets; in P. Dybjer et al. (eds.), Proc. of the Workshop
  on Programming Languages, Marstrand, Sweden, 1987.  

\bibitem[Ja\'s34]{jas}
  S. Ja\'skowski, On the Rules of Suppositon in Formal Logic;
  in: Studia Logica, vol.\ 1, 1934.

\bibitem[Kor09]{ident}
  A. Korni\l owicz,
  {\em How to Define Terms in Mizar Effectively};
  in: Studies in Logic, Grammar and Rhetoric, vol. 18(31), pp. 67-77, 2009.

\bibitem[Mil01a]{complexf} A.J. Milewska,
      The Field of Complex Numbers;
      Formalized Mathematics, 9(2), pp.~265--269, 2001.

\bibitem[Mil01b]{poly} R. Milewski,
      The Ring of Polynomials;
      Formalized Mathematics, 9(2), pp.~339--346, 2001.

\bibitem[Miz10]{HomePage} The Mizar Home Page, {\tt http://mizar.org}.

\bibitem[NB04]{Naumowicz}
  A. Naumowicz and C. Byli\'nski, Improving Mizar texts with properties
  and requirements,
  in: A.~Asperti, G. Bancerek, and A. Trybulec (eds.),
  Proc. of MKM 2004,
  Lecture Notes in Computer Science 3119, pp.~190--301, 2004.

\bibitem[OS98]{oppscha}
  A.V. Oppenheim and R.W. Schafer, Discrete-Time Signal Processing
  (2nd edition); Prenctice-Hall, New Jersey, 1998.

\bibitem[RT01]{mizar}
  P. Rudnicki and A. Trybulec, Mathematical Knowledge Management in Mizar;
  in: B. Buchberger, O. Caprotti (eds.), Proc. of
  MKM 2001, Linz, Austria, 2001.

\bibitem[Sch10]{ratfunc}
  C. Schwarzweller, Rational Functions; to appear in Journal of Formalized Mathematics.

\bibitem[Sch21]{Schur}
 J. Schur, \"{U}ber Algebraische Gleichungen, die nur Wurzeln mit
 negativen Realteilen besitzen;
  in: Zeitschrift f\"{u}r angewandte Mathematik und Mechanik, vol. 1, pp. 95--110, 1921.

\bibitem[SR07]{schur1}
A. Rowinska-Schwarzweller and C. Schwarzweller,
Towards Mathematical Knowledge Management for Electrical Engineering;
in: M. Kauers, M. Kerber, R. Miner, W. Windsteiger (eds.), Towards Mechanized Mathematical Assistants, 
Lecture Notes in Artificial Intelligence 4573, pp. 371-380, 2007.

\bibitem[SR10]{routh}
  C. Schwarzweller and A. Rowinska-Schwarzweller, A Theorem for Checking
  Stability of Networks; to appear in Journal of Formalized Mathematics.

\bibitem[Tar39]{Tarski}
  A. Tarski, On Well-Ordered Subsets of Any Set;
  in: Fundamenta Mathematicae, vol. 32, pp. 176--183, 1939.

\bibitem[Unb93]{unb}
  R. Unbehauen, Netzwerk- und Filtersynthese: Grundlagen und Anwendungen (4. Auflage);
  Oldenbourg-Verlag, 1993.

\end{thebibliography}
\end{document}